\pdfoutput=1

\documentclass[11pt]{article}

\usepackage[final]{acl}

\usepackage{times}
\usepackage{latexsym}

\usepackage[T1]{fontenc}

\usepackage[utf8]{inputenc}

\usepackage{microtype}

\usepackage{inconsolata}

\usepackage{graphicx}

\usepackage{amsmath,amsfonts,bm}

\def\eqref#1{equation~\ref{#1}}

\def\1{\bm{1}}

\newcommand{\test}{\mathcal{D_{\mathrm{test}}}}

\DeclareMathAlphabet{\mathsfit}{\encodingdefault}{\sfdefault}{m}{sl}
\SetMathAlphabet{\mathsfit}{bold}{\encodingdefault}{\sfdefault}{bx}{n}

\newcommand{\E}{\mathbb{E}}

\DeclareMathOperator*{\argmax}{arg\,max}
\DeclareMathOperator*{\argmin}{arg\,min}

\usepackage{hyperref}
\usepackage{url}
\usepackage{enumitem}
\usepackage{algorithm}
\usepackage{algpseudocode}
\usepackage{booktabs}
\usepackage{multirow}
\usepackage{makecell}
\usepackage{xcolor}
\usepackage{colortbl}
\usepackage{subfigure}
\usepackage{comment}

\graphicspath{ {./figs/} }
\DeclareGraphicsExtensions{.png,.pdf}

\algrenewcommand\algorithmicindent{0.4em}

\newcommand{\recall}{\operatorname{ReCaLL}}
\newcommand{\ours}{EM-MIA}
\newcommand{\ourbenchmark}{OLMoMIA}
\newcommand{\lm}{\mathcal{M}}

\newcommand{\rand}{\textit{Rand}}
\newcommand{\randm}{\textit{RandM}}
\newcommand{\randnm}{\textit{RandNM}}
\newcommand{\toppref}{\textit{TopPref}}
\newcommand{\avg}{\textit{Avg}}
\newcommand{\avgp}{\textit{AvgP}}
\newcommand{\argtopk}{\mathop{\mathrm{arg\,topk}}}
\newcommand{\median}{\mathop{\mathrm{median}}}

\definecolor{ReCaLLColor}{gray}{0.85}
\definecolor{OursColor}{rgb}{0.88,1,1}

\title{Detecting Training Data of Large Language Models \\via Expectation Maximization}

\author{{\bf Gyuwan Kim$^{1}$\thanks{ Work done during an internship at AWS AI Labs }} \quad 
{\bf Yang Li$^2$} \quad 
{\bf Evangelia Spiliopoulou$^2$} \quad \\
{\bf Jie Ma$^2$} \quad 
{\bf William Yang Wang$^{1}$\thanks{ Work done while at AWS AI Labs }} \\
$^1$University of California, Santa Barbara \quad $^2$AWS AI Lab \\
\texttt{gyuwankim@ucsb.edu}
}

\begin{document}

\maketitle

\begin{abstract}
Membership inference attacks (MIAs) aim to determine whether a specific example was used to train a given language model.  
While prior work has explored prompt-based attacks such as ReCALL, these methods rely heavily on the assumption that using known non-members as prompts reliably suppresses the model’s responses to non-member queries.  
We propose \ours{}, a new membership inference approach that iteratively refines prefix effectiveness and membership scores using an expectation-maximization strategy without requiring labeled non-member examples.  
To support controlled evaluation, we introduce \ourbenchmark{}, a benchmark that enables analysis of MIA robustness under systematically varied distributional overlap and difficulty.  
Experiments on WikiMIA and \ourbenchmark{} show that \ours{} outperforms existing baselines, particularly in settings with clear distributional separability.  
We highlight scenarios where \ours{} succeeds in practical settings with partial distributional overlap, while failure cases expose fundamental limitations of current MIA methods under near-identical conditions.  
We release our code and evaluation pipeline to encourage reproducible and robust MIA research.

\end{abstract}

\section{Introduction}

As large language models (LLMs)~\citep{brown2020language, touvron2023llama2} continue to advance in scale and capability, growing concerns have emerged regarding the provenance and transparency of their training data~\citep{henderson2023foundation, liang2023holistic}.
This issue is crucial in both research and real-world deployments, where uncertainty about what data a model has seen can lead to legal and ethical risks, such as privacy breaches~\citep{staab2023beyond, kandpal2023user}, copyright infringement~\citep{meeus2024copyright}, and the leakage of sensitive or proprietary content~\citep{chang2023speak}.

Membership inference attacks (MIAs) offer a concrete framework for probing this issue by attempting to determine whether a specific example was included in a model’s training corpus~\citep{shokri2017membership, carlini2022membership}.  
By doing so, they enable auditing of model behavior and exposure, helping practitioners evaluate data contamination~\citep{magar2022data, sainz2023nlp, sainz2024data, ravaut2024comprehensive} or compliance with data usage policies~\citep{voigt2017eu, legislature2018ccpa}.
Despite their utility, MIAs on LLMs remain fundamentally challenging due to the massive size of pre-training corpora and the subtle boundary between memorization and generalization in natural language~\citep{duan2024membership}.  
Recent work has proposed prompt-based MIA techniques such as ReCALL~\citep{xie2024recall}, which assume that known non-members can serve as effective prompts for distinguishing members from non-members.  
However, we find that the effectiveness of such prompts is highly inconsistent and difficult to predict, motivating the need for a more adaptive approach that can account for variability in prompt effectiveness.

To address the limitations of approaches that rely on arbitrarily or randomly chosen prompts, we propose \ours{}, a novel membership inference method that jointly refines prefix effectiveness and membership scores through an expectation-maximization procedure.  
Our approach is motivated by the observation that the usefulness of a prompt, defined as its ability to differentiate members from non-members, varies widely across examples and cannot be reliably determined in advance.  
Instead of relying on labeled non-members or assuming the quality of predefined prompts, \ours{} uses the model’s own responses to iteratively estimate which prefixes are informative and which examples are likely to be members.  
This interaction allows the model to bootstrap its predictions over both prompt selection and membership estimation in a fully unsupervised manner.  
As a result, \ours{} offers greater flexibility and robustness across diverse settings, particularly when prompt-based assumptions do not hold or ground-truth non-member data is unavailable.

To facilitate more controlled and reproducible evaluation of membership inference methods, we introduce \ourbenchmark{}, a benchmark constructed from the pre-training corpus and checkpoints of the OLMo open-source LLM series~\citep{groeneveld2024olmo}.  
Unlike existing benchmarks such as WikiMIA~\citep{shi2023detecting} and MIMIR~\citep{duan2024membership}, which provide limited control over the similarity between member and non-member examples, \ourbenchmark{} allows researchers to systematically vary distributional overlap and assess how different methods perform across a range of difficulty levels.  
By partitioning the data based on semantic similarity and membership status with respect to the pre-training data, \ourbenchmark{} supports fine-grained analysis of robustness, generalization, and failure modes in both easy and near-indistinguishable settings.  
Its design enables rigorous comparison of inference strategies under controlled conditions, and we will release both the benchmark and its generation pipeline to support scalable and reproducible MIA research.

Our experiments show that \ours{} outperforms existing MIA methods on WikiMIA across models of varying sizes and achieves robust results on \ourbenchmark{} under systematically controlled difficulty conditions.  
In particular, \ours{} demonstrates strong performance without access to labeled non-member data and maintains robustness to prompt variability, highlighting its practical value in realistic gray-box scenarios.  
At the same time, our results expose the inherent difficulty of membership inference when member and non-member distributions are nearly identical, which poses a significant challenge for all existing methods, including ours.
These findings underscore the importance of evaluating MIA methods across a range of separability conditions and offer new insight into the limits and opportunities of prompt-based membership inference.

\section{Related Work}
\label{sec:related_work}

\paragraph{Membership Inference on LLMs.}
Membership inference on LLMs presents unique challenges.  
First, LLMs are trained on massive corpora, and individual examples are typically seen only once or a few times~\citep{lee2021deduplicating}, leaving a minimal memorization footprint.  
Second, defining membership is inherently ambiguous in natural language, in that texts often repeat or partially overlap even after rigorous decontamination~\citep{kandpal2022deduplicating, tirumala2024d4}, and paraphrased or semantically similar content can blur membership boundaries~\citep{shilov2024mosaic, mattern2023membership, mozaffari2024semantic}.  
Traditional MIA methods often rely on training shadow models using labeled data from a similar distribution~\citep{shokri2017membership}, but this is impractical in LLM settings due to limited access to comparable data and training specifications.

In contrast, MIA methods for LLMs typically use the model’s loss (e.g., negative log-likelihood) as a membership score, under the assumption that models tend to memorize or overfit their training data~\citep{yeom2018privacy, carlini2022membership}.  
Building on this idea, several techniques calibrate membership scores based on input difficulty~\citep{ye2022enhanced}, using reference models~\citep{carlini2022membership}, compression-based heuristics~\citep{carlini2021extracting}, or nearest neighbors in embedding space~\citep{mattern2023membership}.  
Other methods focus on low-likelihood tokens~\citep{shi2023detecting} or compute calibrated token-level ratios~\citep{zhang2024min}.

ReCALL~\citep{xie2024recall} proposes a different strategy by using known non-member examples as prompts to condition the model's response.  
It assumes that such prompts suppress memorization signals, enabling members to stand out by their elevated likelihood under the same prompt.  
However, this assumption is brittle, as prompt effectiveness varies significantly across examples, and a fixed prompt often fails to generalize across models or domains.  
We address this limitation by proposing a fully unsupervised method that jointly estimates prompt effectiveness and membership likelihood, without relying on labeled non-members or fixed prompting strategies.

\paragraph{Evaluation Benchmarks.}
Robust evaluation of MIA methods for LLMs remains challenging because existing benchmarks rarely provide both reliable membership labels and controllable distributional settings.  
Most benchmarks fall into one of two categories.  
Some, such as WikiMIA~\citep{shi2023detecting, meeus2024did}, determine membership based on document timestamps and model release dates.  
This approach risks conflating membership inference with distribution shift detection~\citep{das2025blind, meeus2024inherent, maini2024llm}.  
Others, such as MIMIR~\citep{duan2024membership}, use random splits to ensure that the distributions of members and non-members are nearly identical.  
In such cases, no existing method performs significantly better than random guessing.

These limitations make it difficult to understand how well a method generalizes across different data conditions.  
Pre-training corpora are typically drawn from diverse sources, while inference-time inputs may come from entirely different domains.  
Effective evaluation, therefore, requires testing under a range of membership separability conditions.  
However, constructing such benchmarks is practically difficult, especially given the lack of true non-member data and the challenge of controlling test distributions.  
There is a clear need for evaluation setups that reflect varied, realistic scenarios while maintaining access to reliable ground-truth labels~\citep{meeus2024inherent, eichler2024nob}.

\section{Method}
\label{sec:method}

\subsection{Problem Formulation}

We consider membership inference in a gray-box setting, where the attacker has access to a language model $\lm$ and can query $\lm$ to obtain token-level probabilities or log-likelihoods.
Given an input $x \in \test$, the goal is to predict a binary membership label indicating whether $x$ was included in the pretraining corpus $\mathcal{D_{\mathrm{train}}}$ of $\lm$.

\subsection{ReCaLL: Assumptions and Limitations}
\label{sec:recall}

ReCaLL~\citep{xie2024recall} is a prompt-based membership inference method that computes the ratio between the conditional and unconditional log-likelihoods of a target example $x$ under $\lm$.
Given a prefix $p$, the ReCaLL score is defined as $\recall_{p}(x; \lm) = \mathrm{LL}(x \mid p; \lm) / \mathrm{LL}(x; \lm)$, where $\mathrm{LL}$ denotes the average log-likelihood over tokens, and $p = p_1 \oplus \cdots \oplus p_n$ is a concatenation of non-member examples $p_i$.
The intuition is that conditioning on non-members tends to reduce the likelihood of members more than that of non-members, making the ratio indicative for membership prediction.

ReCaLL demonstrates strong empirical performance, achieving over 90\% AUC-ROC on WikiMIA~\citep{shi2023detecting} and outperforming prior methods such as Min-K\%++~\citep{zhang2024min}.
However, this performance depends on strong assumptions and lacks theoretical justification.
In its original implementation, ReCaLL constructs prefixes by randomly selecting non-members from the test set, assuming that (1) ground-truth non-members are available at inference time, and (2) all non-members are equally effective as prompts.

In practice, such assumptions rarely hold so labeled non-members are often unavailable, especially when the training and test data distributions substantially overlap~\citep{villalobos2022will, muennighoff2024scaling}.
Even synthetic prefixes generated using GPT-4, as explored in~\citet{xie2024recall}, rely on seed non-members drawn from the test distribution.
This reliance on known non-members gives ReCaLL an unfair advantage over methods that operate without access to test labels.

Ablation studies in~\citet{xie2024recall} further show that ReCaLL's performance degrades when the prefix and test inputs differ in distribution, and that different random samples yield significant variance in accuracy.
These findings suggest that non-members vary widely in their effectiveness as prompts, and that ReCaLL does not generalize reliably across domains or distribution shifts.
These limitations motivate the need for a more flexible and fully unsupervised approach that does not depend on labeled non-members or assume prompt effectiveness in advance.

\begin{figure}[t]
\centering
\includegraphics[width=\linewidth]{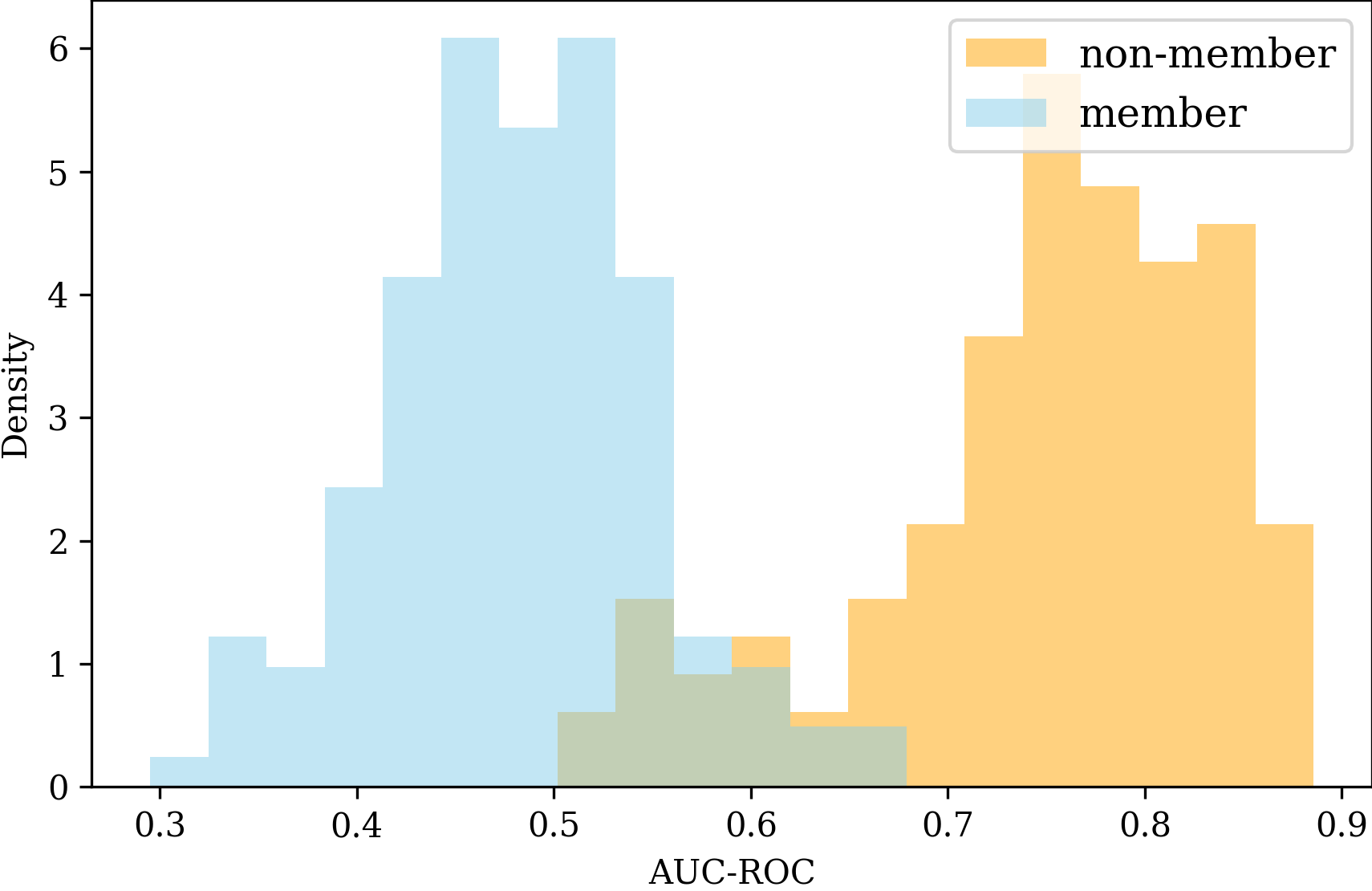}
\caption{
\label{fig:prefix_scores_hist}
Distribution of prefix scores (measured by AUC-ROC in the oracle setting) for members and non-members on WikiMIA~\citep{shi2023detecting} (length 128) using Pythia-6.9B~\citep{biderman2023pythia}.
}
\end{figure}

\begin{figure*}[th!]
\begin{minipage}{.68\textwidth}
\begin{algorithm}[H]
\caption{\ours} \label{alg:recallpp}
\begin{algorithmic}[1]
\Require Target LLM $\mathcal{M}$, Test dataset $\test$
\Ensure Membership scores $f(x)$ for $x \in \test$
\State Initialize $f(x)$ with an existing off-the-shelf MIA method
\Repeat
\State Update prefix scores $r(p) = S(\recall_{p}, f, \test)$ for $p \in \test $ \par
\State Update membership scores $f(x) = -r(x)$ for $x \in \test $ \par 
\Until{Convergence (no significant difference in $f$)}
\end{algorithmic}
\end{algorithm}
\end{minipage}
\hfill
\begin{minipage}{.30\textwidth}
\includegraphics[width=\textwidth]{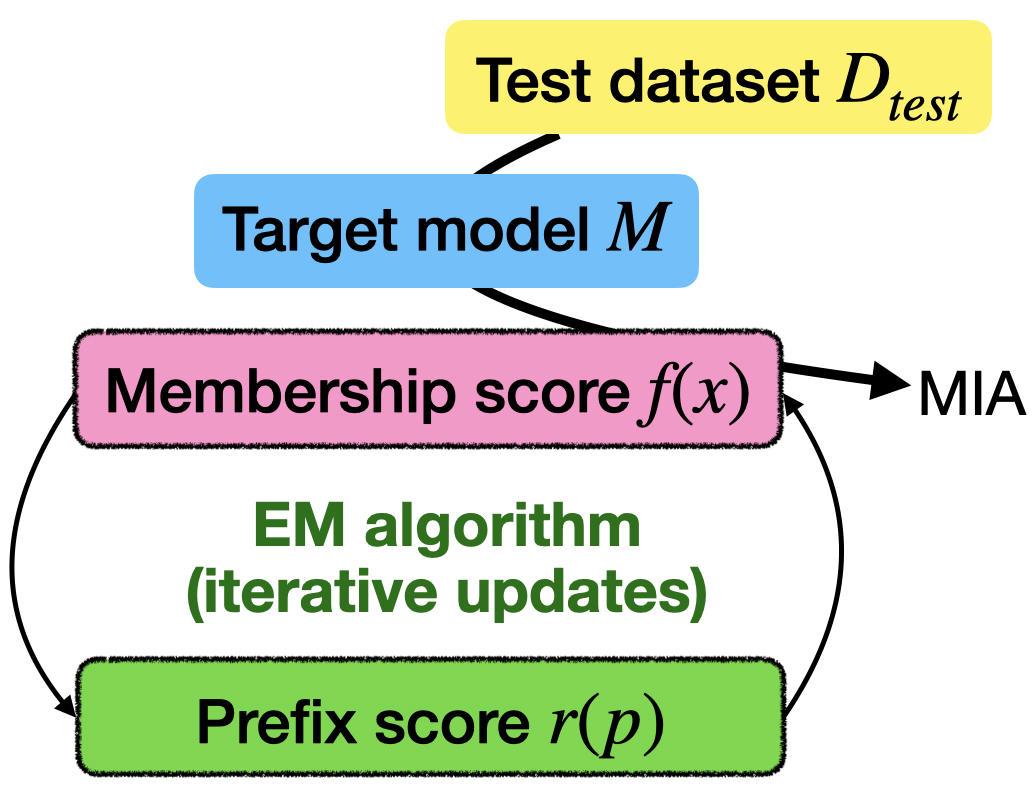}
\end{minipage}
\end{figure*}

\subsection{Motivation: Sensitivity to Prefix Choice}
\label{sec:observation}

We empirically examine how ReCaLL’s performance varies with the choice of prefix, particularly when labeled non-members are unavailable.
To this end, we define a \emph{prefix score} $r(p)$ as the effectiveness of a prefix $p$ in distinguishing members from non-members when used in ReCaLL.

In an oracle setting with access to ground-truth membership labels, we compute $r(p)$ as the AUC-ROC of $\recall_p(x)$ over a test set $\test$, using each $x \in \test$ as a standalone prefix.
This allows us to empirically measure the effectiveness of each test example when used as a prefix.

Figure~\ref{fig:prefix_scores_hist} shows that non-member prefixes generally lead to strong ReCaLL performance, with AUC-ROC often exceeding 0.7.  
In contrast, member prefixes perform poorly, with scores clustering near 0.5 (i.e., random guessing).  
Additional comparisons using alternative metrics for prefix scoring are included in Appendix~\ref{sec:prefix_scores_metrics}.
These results highlight two limitations of current ReCaLL-based methods:  
(1) Even among non-members, prefix effectiveness varies widely;   
(2) In realistic scenarios, ground-truth labels needed to evaluate or filter prefixes are unavailable.  

These findings underscore the need for an approach that can identify effective prefixes and infer membership without access to labels.
We address this challenge in the following section by proposing a fully unsupervised method that jointly estimates membership likelihood and prefix effectiveness through iterative refinement.

\subsection{EM-MIA: Joint Estimation via EM}
\label{sec:em-mia}

To address the practical setting where neither labeled non-members nor reliable prompt effectiveness can be assumed, we propose \ours{}, a fully unsupervised method that jointly estimates prefix effectiveness and membership likelihood using an expectation-maximization (EM) procedure.

Let $f(x)$ denote the membership score for each test example $x \in \test$, and $r(p)$ denote the effectiveness score of a prefix $p$.
The key insight is that membership scores and prefix scores can reinforce each other: better membership estimates allow more accurate estimation of prefix effectiveness, and more reliable prefixes lead to improved membership predictions.
This mutual dependency motivates an iterative procedure in which each set of scores is refined based on the other.

Algorithm~\ref{alg:recallpp} outlines the overall procedure of \ours.
We initialize membership scores using any existing off-the-shelf MIA method such as Loss~\citep{yeom2018privacy} or Min-K\%++~\citep{zhang2024min} (Line 1).
We then alternate between two updates: (1) estimating prefix scores $r(p)$ based on current membership scores $f(x)$ (Line 3), and (2) updating $f(x)$ using the refined $r(p)$ (Line 4).
This process continues until convergence (Line 5).
As \ours{} is a general framework, initialization, score update rules, stopping criteria, and datasets can be adapted to different applications.

\paragraph{Using External Data.}
Our framework can be extended by incorporating external data to provide additional signals for membership inference. 
Suppose we have a set of known members $\mathcal{D_{\mathrm{m}}}$, known non-members $\mathcal{D_{\mathrm{nm}}}$, and instances with unknown membership $\mathcal{D_{\mathrm{unk}}}$. 
For example, $\mathcal{D_{\mathrm{m}}}$ may consist of older Wikipedia documents, which are commonly included in LLM training corpora. 
As discussed above, we focus on the setting where $\mathcal{D_{\mathrm{nm}}} = \emptyset$, or at least $\mathcal{D_{\mathrm{nm}}} \cap \test = \emptyset$, although non-members could also be constructed from clearly unnatural texts (e.g., ``*b9qx84;5zln''). 
Ideally, $\mathcal{D_{\mathrm{unk}}}$ is drawn from the same distribution as $\test$, but it may come from any corpus when the test distribution is unknown. 
All available data can then be incorporated to improve the estimation of membership and prefix scores by augmenting the test set as $\test \leftarrow \test \cup \mathcal{D_{\mathrm{m}}} \cup \mathcal{D_{\mathrm{nm}}} \cup \mathcal{D_{\mathrm{unk}}}$.
We leave a systematic study of this extension for future work.

\paragraph{Updating Prefix Scores.}
As shown in Section~\ref{sec:observation}, AUC-ROC is an effective function $S$ for evaluating a prefix $p$ in the oracle setting given ground truth labels.
Since ground-truth labels are not available, we generate pseudo-labels using a threshold $\tau$ over current membership scores $f(x)$ and use them to calculate prefix scores: $\text{AUC-ROC}(\{(\recall_p(x), \mathbf{1}_{f(x) > \tau}) \mid x \in \test \})$.
We typically set $\tau$ to the median of $f(x)$, assuming a balanced dataset.
Alternatively, instead of relying on hard thresholds, we can measure rank alignment between $\recall_p(x)$ and $f(x)$ using the average absolute rank difference or rank correlation coefficients such as Kendall's tau~\citep{kendall1938new} or Spearman’s rho~\citep{spearman1961proof}.

\paragraph{Updating Membership Scores.}
Section~\ref{sec:observation} also shows that a negative prefix score $-r(x)$ is a simple yet effective membership score.
Alternatively, one could construct a prefix $p = p_1 \oplus \cdots \oplus p_n$ using top-$k$ examples ranked by $r(x)$, and compute $f(x) = \recall_p(x)$ using this prefix.
The ordering of $p_i$ within $p$ is also a design choice. 
Placing stronger prefixes closer to $x$ may amplify their influence due to LLMs’ attention bias toward recent tokens.

\begin{figure}[t!]
\centering

\includegraphics[width=\linewidth]{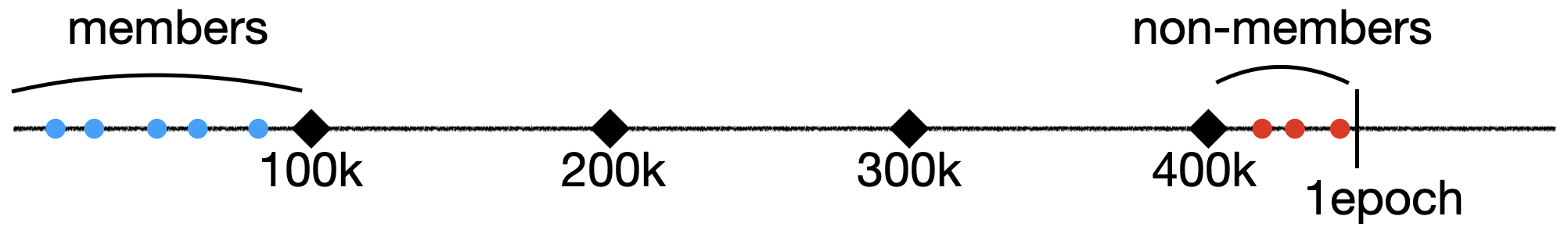}

\caption{The basic setup of \ourbenchmark~benchmark. The horizontal line indicates a training step. For any intermediate checkpoint at a specific step, we can consider training data before and after that step as members and non-members, respectively.
\label{fig:olmomia}
}
\end{figure}

\section{OLMoMIA Benchmark}
\label{sec:olmomia}

\paragraph{Motivation.}
To enable controlled and reproducible evaluation of MIA methods under varying difficulty levels, we introduce \ourbenchmark{}, a new benchmark constructed from the training data and checkpoints of the OLMo-7B model~\citep{groeneveld2024olmo}, which was pre-trained on the Dolma dataset~\citep{soldaini2024dolma}.  
Unlike existing benchmarks such as WikiMIA~\citep{shi2023detecting}, which rely on time-based heuristics, or MIMIR~\citep{duan2024membership}, which draws member and non-member examples from randomly partitioned subsets of the same data distribution, \ourbenchmark{} allows systematic control over the distributional overlap between members and non-members.  
This allows evaluation under more realistic and ambiguous conditions, where membership inference is inherently more difficult.

\paragraph{Membership Label Assignment.}
Figure~\ref{fig:olmomia} illustrates the benchmark setup.
OLMo provides intermediate model checkpoints and a detailed index mapping training steps to data examples, offering a rare opportunity to precisely define membership.  
We use four OLMo-7B checkpoints saved at 100k, 200k, 300k, and 400k training steps, where one full epoch consists of just over 450k steps.  
We define member examples as those seen before step 100k and non-members as those introduced between steps 400k and 500k.  
This setup reflects a practical incremental training scenario.  
Some ambiguity in membership may remain despite deduplication, as discussed in Section~\ref{sec:related_work}.

\paragraph{Dataset Sampling with Varying Difficulty.}
We construct six dataset variants to simulate different levels of distributional overlap between members and non-members.
The \textit{Random} setting samples member and non-member examples uniformly, without explicitly accounting for their relative similarity, forming $\mathcal{D_{\mathrm{random}}} = \mathcal{D^\mathrm{m}_{\mathrm{random}}} \cup \mathcal{D^\mathrm{nm}_{\mathrm{random}}}$.
This setting is analogous to MIMIR~\citep{duan2024membership}, which is known to be more challenging than WikiMIA due to minimal distributional differences between members and non-members~\citep{gao2020pile}.

To introduce controlled variation in difficulty, we embed all candidate examples using NV-Embed-v2~\citep{lee2024nv}, the top-performing model on the MTEB leaderboard~\citep{muennighoff2022mteb} as of August 2024.  
We then perform K-means clustering~\citep{lloyd1982least} separately on member and non-member embeddings with $K = 50$.  
To ensure sufficient diversity within clusters, we apply greedy deduplication by removing examples whose cosine distance to another point in the same cluster is below $0.6$.
After this filtering step, we obtain $K$ member clusters $\{C^{m}_i\}_{i=1}^K$ and $K$ non-member clusters $\{C^{nm}_j\}_{j=1}^K$, where these clusters satisfy $d(x, y) > 0.6$ for all $x, y \in C^{m}_i$ and $d(x, y) > 0.6$ for all $x, y \in C^{nm}_j$. 

Based on these clusters, we construct difficulty-controlled variants of \ourbenchmark{} by selecting a member-non-member cluster pair $(i,j)$ and sampling
examples according to their inter-cluster separability.
For any pair $(i,j)$, we measure inter-cluster distance as $\Delta(i,j) = \E_{x \in C^{m}_i,\, y \in C^{nm}_j} d(x,y)$.

In the \textit{Easy} setting, we maximize separability by first selecting
the most dissimilar member--non-member cluster pair $(i_{\mathrm{easy}}, j_{\mathrm{easy}}) = \argmax_{(i,j)} \Delta(i,j)$ and then choosing the member examples that are, on average, farthest from the selected non-member cluster and the non-member examples farthest from the member cluster, which we formalize as $\mathcal{D^\mathrm{m}_{\mathrm{easy}}} = \argtopk_{x \in C^{m}_{i_{\mathrm{easy}}}} \E_{y \in C^{nm}_{j_{\mathrm{easy}}}} d(x,y)$ and $\mathcal{D^\mathrm{nm}_{\mathrm{easy}}} = \argtopk_{y \in C^{nm}_{j_{\mathrm{easy}}}} \E_{x \in C^{m}_{i_{\mathrm{easy}}}} d(x,y)$, yielding $\mathcal{D_{\mathrm{easy}}} = \mathcal{D^\mathrm{m}_{\mathrm{easy}}} \cup \mathcal{D^\mathrm{nm}_{\mathrm{easy}}}$.

In the \textit{Hard} setting, we minimize separability by first selecting the most similar member--non-member cluster pair $(i_{\mathrm{hard}}, j_{\mathrm{hard}}) = \argmin_{(i,j)} \Delta(i,j)$ and then choosing the member and non-member examples that are, on average, closest to the opposing cluster, which we formalize as $\mathcal{D^\mathrm{m}_{\mathrm{hard}}} = \argtopk_{x \in C^{m}_{i_{\mathrm{hard}}}} \bigl(-\E_{y \in C^{nm}_{j_{\mathrm{hard}}}} d(x,y)\bigr)$ and $\mathcal{D^\mathrm{nm}_{\mathrm{hard}}} = \argtopk_{y \in C^{nm}_{j_{\mathrm{hard}}}} \bigl(-\E_{x \in C^{m}_{i_{\mathrm{hard}}}} d(x,y)\bigr)$, yielding $\mathcal{D_{\mathrm{hard}}} = \mathcal{D^\mathrm{m}_{\mathrm{hard}}} \cup \mathcal{D^\mathrm{nm}_{\mathrm{hard}}}$.

The \textit{Medium} setting represents intermediate difficulty by selecting the member--non-member cluster pair whose inter-cluster distance is the median over all pairs, $(i_{\mathrm{medium}}, j_{\mathrm{medium}}) = \median_{(i,j)} \Delta(i,j)$, and then randomly selecting member and non-member examples from the corresponding clusters, yielding $\mathcal{D_{\mathrm{medium}}} = \mathcal{D^\mathrm{m}_{\mathrm{medium}}} \cup \mathcal{D^\mathrm{nm}_{\mathrm{medium}}}$, where $\mathcal{D^\mathrm{m}_{\mathrm{medium}}} \subset C^{m}_{i_{medium}}$ and $\mathcal{D^\mathrm{nm}_{\mathrm{medium}}} \subset C^{nm}_{j_{medium}}$.

Finally, we define two hybrid settings that decouple the difficulty of member and non-member distributions.
\textit{Mix-1} combines randomly selected members with hard non-members, $\mathcal{D_{\mathrm{mix\text{-}1}}} = \mathcal{D^\mathrm{m}_{\mathrm{random}}} \cup \mathcal{D^\mathrm{nm}_{\mathrm{hard}}}$, while \textit{Mix-2} combines hard members with randomly selected non-members, $\mathcal{D_{\mathrm{mix\text{-}2}}} = \mathcal{D^\mathrm{m}_{\mathrm{hard}}} \cup \mathcal{D^\mathrm{nm}_{\mathrm{random}}}$.
Together, these configurations span a broad range of separability
conditions and provide a controlled testbed for evaluating membership
inference attacks.

\paragraph{Dataset Specifications.}
Each difficulty variant includes two subsets with maximum sequence lengths of 64 and 128 tokens.  
Each subset contains 500 members and 500 non-members, for a total of 1,000 examples per dataset.

\section{Experimental Setup}

Our implementation and datasets to reproduce our experiments are available at \href{https://github.com/gyuwankim/em-mia}{https://github.com/gyuwankim/em-mia}.

\subsection{Datasets and Models}
We evaluate \ours{} and compare it with baseline methods on WikiMIA (\S\ref{sec:results-wikimia}) and \ourbenchmark{} (\S\ref{sec:results-olmomia}) using AUC-ROC as a main evaluation metric.
We also report TPR@1\%FPR results in Appendix~\ref{sec:tpr_at_low_fpr}.
WikiMIA~\citep{shi2023detecting} provides length-based splits of 32, 64, and 128, and we follow prior work~\citep{xie2024recall, zhang2024min} in using Mamba 1.4B~\citep{gu2023mamba}, Pythia 6.9B~\citep{biderman2023pythia}, GPT-NeoX 20B~\citep{black2022gpt}, LLaMA 13B/30B~\citep{touvron2023llama}, and OPT 66B~\citep{zhang2022opt} as target models.
For \ourbenchmark{}, we use all six controlled difficulty settings of \textit{Easy}, \textit{Medium}, \textit{Hard}, \textit{Random}, \textit{Mix-1}, and \textit{Mix-2}, and evaluate using OLMo-7B checkpoints after 100k, 200k, 300k, and 400k training steps.  
We exclude MIMIR~\citep{duan2024membership} from our experiments since it lacks a baseline that performs meaningfully better than random guessing, which is required for initialization in \ours{}.

\subsection{Baselines}
\label{sec:experiments-baselines}
We compare \ours~against the following baselines: Loss~\citep{yeom2018privacy}, Ref~\citep{carlini2022membership}, Zlib~\citep{carlini2021extracting}, Min-K\%~\citep{shi2023detecting}, and Min-K\%++~\citep{zhang2024min}.  
We use Pythia-70m for WikiMIA and StableLM-Base-Alpha-3B-v2 model~\citep{StableLMAlphaV2Models} for \ourbenchmark~as the reference model of the Ref method, following~\citet{shi2023detecting} and \citet{duan2024membership}.
We use $K = 20$ for Min-K\% and Min-K\%++.
Among the commonly used baselines, we omit Neighbor~\citep{mattern2023membership} because it is not the best in most cases though it requires LLM inference multiple times for neighborhood texts, so it is much more expensive.

\subsection{ReCaLL-based Baselines}
We include several variants of ReCaLL that differ in how the prefix $p = p_1 \oplus \cdots \oplus p_n$ is constructed: \rand, \randm, \randnm, and \toppref.
\rand~randomly selects any data from $\test$.
\randm~randomly selects member data from $\test$.
\randnm~randomly selects non-member data from $\test$.
\toppref~selects data from $\test$ with the highest prefix scores calculated with ground truth labels the same as \S\ref{sec:observation}.

Among these, only \rand is fully unsupervised; the others either partially or fully rely on labels in the test dataset, making them unsuitable for realistic scenarios.  
For all methods using a random selection (\rand, \randm, and, \randnm), we execute five times with different random seeds and report the average.
We fix $n = 12$ since it provides a reasonable performance while not too expensive.
We report the results from the original ReCaLL paper but explain why this is not a fair comparison in Appendix~\ref{sec:recall_unfairness}.

We also evaluate two unsupervised averaging variants. \textit{Avg} and \textit{AvgP} average ReCaLL scores over all data points in $\test$: $\textit{Avg}(x) = \frac{1}{|\test|} \sum_{p \in \test} \recall_p(x)$ and $\textit{AvgP}(p) = \frac{1}{|\test|} \sum_{x \in \test} \recall_p(x)$. 
The intuition is averaging will smooth out ReCaLL scores with a non-discriminative prefix while keeping ReCaLL scores with a discriminative prefix without exactly knowing discriminative prefixes.

\begin{table*}[!th]

\begin{center}
\setlength\tabcolsep{2pt}
\fontsize{8}{9.5}\selectfont

\begin{tabular}{lccccccccccccccccccccc}
\toprule
{\bf  Method} & \multicolumn{3}{c}{\bf  \makecell{Mamba-1.4B}} & \multicolumn{3}{c}{\bf  \makecell{Pythia-6.9B}} & \multicolumn{3}{c}{\bf  \makecell{LLaMA-13B}} & \multicolumn{3}{c}{\bf  \makecell{NeoX-20B}} & \multicolumn{3}{c}{\bf  \makecell{LLaMA-30B}} & \multicolumn{3}{c}{\bf  \makecell{OPT-66B}} & \multicolumn{3}{c}{\bf  Average} \\ 
\cmidrule(lr){2-4} \cmidrule{5-7} \cmidrule(lr){8-10} \cmidrule(lr){11-13} \cmidrule(lr){14-16} \cmidrule(lr){17-19} \cmidrule(lr){20-22}
{\bf } & 32 & 64 & 128 & 32 & 64 & 128 & 32 & 64 & 128 & 32 & 64 & 128 & 32 & 64 & 128 & 32 & 64 & 128 & 32 & 64 & 128 \\
\midrule
Loss      & 61.0 & 58.2 & 63.3 & 63.8 & 60.8 & 65.1 & 67.5 & 63.6 & 67.7 & 69.1 & 66.6 & 70.8 & 69.4 & 66.1 & 70.3 & 65.7 & 62.3 & 65.5 & 66.1 & 62.9 & 67.1 \\
Ref       & 60.3 & 59.7 & 59.7 & 63.2 & 62.3 & 63.0 & 64.0 & 62.5 & 64.1 & 68.2 & 67.8 & 68.9 & 65.1 & 64.8 & 66.8 & 63.9 & 62.9 & 62.7 & 64.1 & 63.3 & 64.2 \\
Zlib      & 61.9 & 60.4 & 65.6 & 64.3 & 62.6 & 67.6 & 67.8 & 65.3 & 69.7 & 69.3 & 68.1 & 72.4 & 69.8 & 67.4 & 71.8 & 65.8 & 63.9 & 67.4 & 66.5 & 64.6 & 69.1 \\
Min-K\%   & 63.3 & 61.7 & 66.7 & 66.3 & 65.0 & 69.5 & 66.8 & 66.0 & 71.5 & 72.1 & 72.1 & 75.7 & 69.3 & 68.4 & 73.7 & 67.5 & 66.5 & 70.6 & 67.5 & 66.6 & 71.3 \\
Min-K\%++ & 66.4 & 67.2 & 67.7 & 70.2 & 71.8 & 69.8 & 84.4 & 84.3 & 83.8 & 75.1 & 76.4 & 75.5 & 84.3 & 84.2 & 82.8 & 69.7 & 69.8 & 71.1 & 75.0 & 75.6 & 75.1 \\
\cmidrule{1-22}
\rowcolor{ReCaLLColor}
\avg      & 70.2 & 68.3 & 65.6 & 69.3 & 68.2 & 66.7 & 77.2 & 77.3 & 74.6 & 71.4 & 72.0 & 68.7 & 79.8 & 81.0 & 79.6 & 64.6 & 65.6 & 60.0 & 72.1 & 72.1 & 69.2 \\
\rowcolor{ReCaLLColor}
\avgp     & 64.0 & 61.8 & 56.7 & 62.1 & 61.0 & 59.0 & 63.1 & 60.3 & 56.4 & 63.9 & 61.8 & 61.1 & 60.3 & 60.0 & 55.4 & 86.9 & 94.3 & 95.1 & 66.7 & 66.5 & 63.9 \\
\rowcolor{ReCaLLColor}
\randm    & 25.4 & 25.1 & 26.2 & 24.9 & 26.2 & 24.6 & 21.0 & 14.9 & 68.6 & 25.3 & 28.3 & 29.8 & 14.0 & 15.1 & 70.4 & 33.9 & 40.9 & 42.9 & 24.1 & 25.1 & 43.8 \\
\rowcolor{ReCaLLColor}
\rand      & 72.7 & 78.2 & 64.2 & 67.0 & 73.4 & 68.7 & 73.9 & 75.4 & 68.5 & 68.2 & 74.5 & 67.5 & 66.9 & 71.7 & 70.2 & 64.5 & 67.8 & 58.6 & 68.9 & 73.5 & 66.3 \\
\rowcolor{ReCaLLColor}
\randnm   & 90.7 & 90.6 & 88.4 & 87.3 & 90.0 & 88.9 & 92.1 & 93.4 & 68.8 & 85.9 & 89.9 & 86.3 & 90.6 & 92.1 & 71.8 & 78.7 & 77.6 & 67.8 & 87.5 & 88.9 & 78.7 \\
\rowcolor{ReCaLLColor}
\toppref   & 90.6 & 91.2 & 88.0 & 91.3 & 92.9 & 90.1 & 93.5 & 94.2 & 71.8 & 88.4 & 92.0 & 90.2 & 92.9 & 93.8 & 74.8 & 83.6 & 79.6 & 72.1 & 90.0 & 90.6 & 81.2 \\
\rowcolor{ReCaLLColor}
\citet{xie2024recall} & 90.2 & 91.4 & 91.2 & 91.6 & 93.0 & 92.6 & 92.2 & 95.2 & 92.5 & 90.5 & 93.2 & 91.7 & 90.7 & 94.9 & 91.2 & 85.1 & 79.9 & 81.0 & 90.1 & 91.3 & 90.0 \\
\cmidrule{1-22}
\rowcolor{OursColor}
\ours     & 97.1 & 97.6 & 96.8 & 97.5 & 97.5 & 96.4 & 98.1 & 98.8 & 97.0 & 96.1 & 97.6 & 96.3 & 98.5 & 98.8 & 98.5 & 99.0 & 99.0 & 96.7 & 97.7 & 98.2 & 96.9 \\
\bottomrule
\end{tabular}
\vspace{-0.5em}
\end{center}

\caption{
\label{tab:wikimia}AUC-ROC results on WikiMIA benchmark. The second block (grey) is the ReCaLL-based baselines. \randm, \randnm, ReCaLL, and \toppref~use labels in the test dataset, so comparing them with others is unfair. We report their scores for reference. We borrow the original ReCaLL results from \citet{xie2024recall}, which is also unfair to be compared with ours and other baselines.
}

\end{table*}

\subsection{\ours}
As described in Section~\ref{sec:em-mia}, \ours{} is a general framework where each component can be tuned for improvement, but we use the following options as defaults based on results from preliminary experiments. 
Overall, Min-K\%++ performs best among baselines without ReCaLL-based approaches, so we use it as a default choice for initialization.
Alternatively, we may use ReCaLL-based methods that do not rely on any labels like \avg, \avgp, or \rand.
For the update rule for prefix scores, we use AUC-ROC as a default scoring function $S$.
For the update rule for membership scores, we use negative prefix scores as new membership scores.
For the stopping criterion, we repeat ten iterations and stop without thresholding by the score difference since we observed that membership scores and prefix scores converge quickly after a few iterations.
We also observed that \ours{} is not sensitive to the choice of the initialization method and the scoring function $S$ and converges to similar results.
Ablation study on different initializations and scoring functions can be found in Section~\ref{sec:iteration}.
Discussion on computational costs can be found in Appendix~\ref{sec:computational_costs}.

\section{Results and Analysis}

\subsection{WikiMIA}
\label{sec:results-wikimia}

Table~\ref{tab:wikimia} and Table~\ref{tab:wikimia_tpr_at_low_fpr} show results on WikiMIA, using AUC-ROC and TPR@1\%FPR as evaluation metrics, respectively.
\ours{} achieves state-of-the-art performance across all models and length splits, significantly outperforming all baselines, including ReCaLL, even without access to labeled non-member examples.  
In all cases, \ours{} exceeds 96\% AUC-ROC.
For the largest model, OPT-66B, it achieves over 99\% AUC-ROC at lengths 32 and 64, whereas ReCaLL falls below 86\%.

\begin{table*}[!th]
\begin{center}
\fontsize{8}{9.5}\selectfont

\begin{tabular}{lcccccccccccc}
\toprule
{\bf  Method} & \multicolumn{2}{c}{\bf  \makecell{Easy}} & \multicolumn{2}{c}{\bf  \makecell{Medium}} & \multicolumn{2}{c}{\bf  \makecell{Hard}} & \multicolumn{2}{c}{\bf  \makecell{Random}} & \multicolumn{2}{c}{\bf  \makecell{Mix-1}} & \multicolumn{2}{c}{\bf  \makecell{Mix-2}} \\ 
\cmidrule(lr){2-3} \cmidrule{4-5} \cmidrule(lr){6-7} \cmidrule(lr){8-9} \cmidrule(lr){10-11} \cmidrule(lr){12-13}
{\bf } & 64 & 128 & 64 & 128 & 64 & 128 & 64 & 128 & 64 & 128 & 64 & 128 \\
\midrule
Loss      & 32.5 & 63.3 & 58.9 & 49.0 & 43.3 & 51.5 & 51.2 & 52.3 & 65.7 & 49.0 & 30.8 & 54.7 \\
Ref       & 56.8 & 26.8 & 61.4 & 47.2 & 49.1 & 50.7 & 49.7 & 49.9 & 59.9 & 49.7 & 38.9 & 50.9 \\
Zlib      & 24.0 & 51.8 & 44.8 & 50.7 & 40.5 & 51.1 & 52.3 & 50.5 & 63.2 & 47.2 & 31.5 & 54.3 \\
Min-K\%   & 32.4 & 50.0 & 54.0 & 51.9 & 43.0 & 51.2 & 51.7 & 51.0 & 60.8 & 50.4 & 34.9 & 51.7 \\
Min-K\%++ & 45.2 & 59.4 & 56.4 & 45.7 & 46.4 & 51.4 & 51.0 & 51.9 & 57.9 & 50.0 & 39.8 & 53.2 \\
\cmidrule{1-13}
\rowcolor{ReCaLLColor}
\avg      & 61.9 & 53.9 & 52.3 & 57.0 & 47.6 & 51.5 & 50.3 & 48.6 & 63.3 & 56.4 & 35.5 & 44.4 \\
\rowcolor{ReCaLLColor}
\avgp     & 79.2 & 39.9 & 53.9 & 61.7 & 50.2 & 51.4 & 49.0 & 50.1 & 55.7 & 63.0 & 42.7 & 41.8 \\
\rowcolor{ReCaLLColor}
\randm    & 32.3 & 22.7 & 39.2 & 30.3 & 45.8 & 50.5 & 48.1 & 48.2 & 49.7 & 48.0 & 29.1 & 28.7 \\
\rowcolor{ReCaLLColor}
\rand     & 63.7 & 46.3 & 56.0 & 59.4 & 48.9 & 52.1 & 49.7 & 49.1 & 60.6 & 68.0 & 38.0 & 38.6 \\
\rowcolor{ReCaLLColor}
\randnm   & 87.1 & 75.5 & 71.8 & 81.2 & 50.5 & 53.2 & 50.4 & 50.0 & 66.5 & 73.7 & 49.1 & 48.0 \\
\rowcolor{ReCaLLColor}
\toppref  & 88.9 & 88.5 & 79.7 & 64.4 & 55.7 & 54.5 & 52.3 & 52.7 & 79.9 & 80.2 & 55.3 & 62.1 \\
\cmidrule{1-13}
\rowcolor{OursColor}
\ours     & 99.8 & 97.4 & 98.3 & 99.8 & 47.2 & 50.2 & 51.4 & 50.9 & 88.3 & 80.8 & 88.4 & 77.1 \\
\bottomrule
\end{tabular}
\vspace{-0.5em}
\end{center}

\caption{
\label{tab:olmomia}AUC-ROC results on \ourbenchmark~benchmark. The second block (grey) is the ReCaLL-based baselines. \randm, \randnm, ReCaLL, and \toppref~use labels in the test dataset, so comparing them with others is unfair. We report their scores for reference.
}

\end{table*}

All non-ReCaLL baselines remain below 76\% AUC-ROC on average.
The performance order among ReCaLL-based variants is consistent: \randm{} < \avg{}, \avgp{} < \rand{} < \randnm{} < \toppref{}.
This pattern confirms that ReCaLL is highly sensitive to the choice of prefix.
Particularly, the significant performance gap between \rand{} and \randnm{} highlights ReCaLL’s reliance on the availability of given non-members.
Importantly, \rand{}, which uses no test labels, performs worse than Min-K\%++ on average, indicating that ReCaLL alone is insufficient under a fully unsupervised setting.

\randnm{} is similar to the original ReCaLL~\citep{xie2024recall} in most cases except for the OPT-66B model and LLaMA models with sequence length 128, probably because $n = 12$ is not optimal for these cases.
\toppref{} consistently outperforms \randnm{}, demonstrating that prefix quality varies and that random prefix selection is suboptimal.
This opens the door to prefix optimization~\citep{shin2020autoprompt, deng2022rlprompt, guo2023connecting}, though finding high-quality prefixes without supervision remains challenging.
Our method approximates prefix quality without labels and uses it to improve membership prediction.

\subsection{\ourbenchmark}
\label{sec:results-olmomia}

Table~\ref{tab:olmomia} and Table~\ref{tab:olmomia_tpr_at_low_fpr} show results on \ourbenchmark, using AUC-ROC and TPR@1\%FPR as evaluation metrics, respectively.
\ours{} performs nearly perfectly on \textit{Easy} and \textit{Medium}, similar to its performance on WikiMIA.
We did not observe consistent differences across checkpoints, despite the expectation that earlier training data would be harder to detect. Therefore, we report averages across four OLMo checkpoints. 
In contrast, it performs close to random guessing on \textit{Hard} and \textit{Random}, similar to MIMIR, where member and non-member distributions heavily overlap and all methods are not sufficiently better than random guessing.
On \textit{Mix-1} and \textit{Mix-2}, \ours{} achieves reasonable scores, though not as high as in easier settings.
In all but the hardest scenarios, \ours{} significantly outperforms all baselines.

None of the baselines without ReCaLL-based approaches are successful in all settings, which implies that \ourbenchmark{} is a challenging benchmark.
The relative order between ReCaLL-based baselines is again consistent: \randm{} < \avg{}, \avgp{}, \rand{} < \randnm{} < \toppref{}, although none of the fully unsupervised variants are successful overall.

Interestingly, \randnm{} works reasonably well on \textit{Mix-1} but does not work well on \textit{Mix-2}.
This is likely because non-members from \textit{Mix-1} are from the same cluster, while non-members from \textit{Mix-1} are randomly sampled from the entire distribution.
\toppref{} again outperforms \randnm{}, reinforcing that not all non-members are equally effective as prompts.

Evaluating MIA for LLMs is difficult because test-time data distributions are unknown.
Benchmarks like \ourbenchmark{} that simulate varied scenarios offer a more comprehensive lens than fixed-split benchmarks.
We encourage future work to assess methods across multiple difficulty levels.
While \ourbenchmark{} is not intended as a strictly more realistic benchmark, it captures plausible conditions not reflected in prior datasets.
Our results show that \ours{} maintains strong performance across a wide spectrum of distributional overlap.

\subsection{Ablation Study on Initializations and Scoring Functions}
\label{sec:iteration}

\begin{figure}[t!]
\centering

\includegraphics[width=\linewidth]{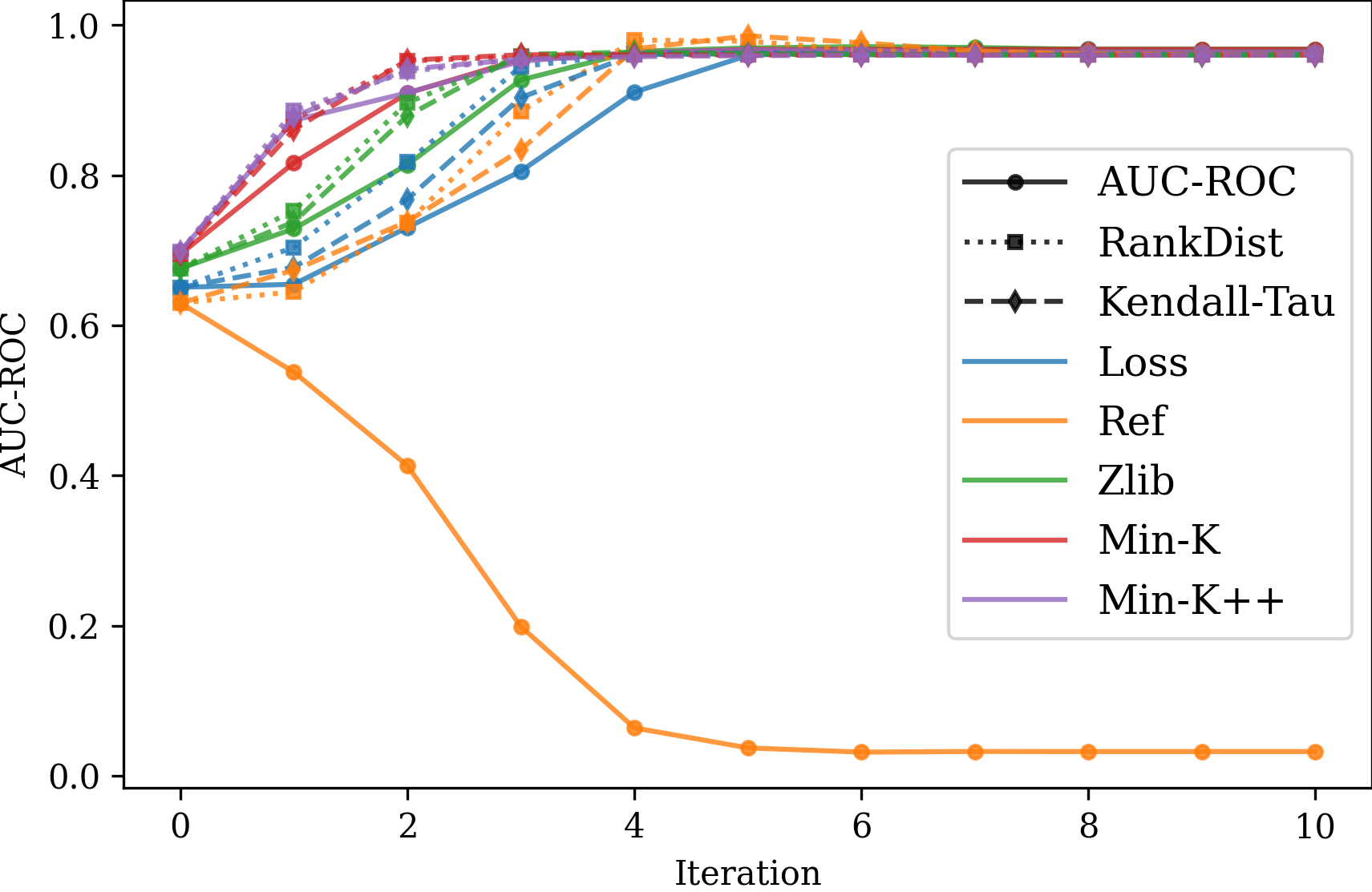}

\caption{Performance of \ours~for each iteration with varying baselines for initialization and scoring functions $S$ on WikiMIA~\citep{shi2023detecting} (length 128) using Pythia-6.9B~\citep{biderman2023pythia}.
\label{fig:iteration}
}
\end{figure}

Figure~\ref{fig:iteration} shows the ablation study on initialization methods (Loss, Ref, Zlib, Min-K\%, Min-K\%++) and prefix scoring functions (AUC-ROC, RankDist, and Kendall-Tau), using WikiMIA with length 128 and Pythia-6.9B.
Each curve shows the change in AUC-ROC calculated from membership scores at each iteration of the expectation-maximization algorithm.
In most combinations, \ours{} converges to a similar accuracy within 4–5 iterations.
In this figure, there is only one case in which the AUC-ROC decreases rapidly and approaches 0.
It is difficult to know when this happens, but it predicts members and non-members in opposite directions, meaning that negative membership scores yield a good AUC-ROC.

\section{Discussion on Computational Costs}
\label{sec:computational_costs}

MIAs for LLMs rely solely on inference without any additional training, so accuracy is typically prioritized over computational costs as long as it remains practically feasible.
In our experiments on WikiMIA and OLMoMIA, all methods were computationally manageable in an academic setting. 
Here, we compare the computational complexity of \ours{} and other baselines, primarily ReCaLL, and describe how the computational costs of \ours{} can be further reduced below.

\ours{} is a general framework whose update rules for prefix scores and membership scores can be flexibly designed (see \S\ref{sec:method}), leading to different trade-offs between accuracy and efficiency. 
For the design used in our experiments (Algorithm~\ref{alg:recallpp}), EM-MIA computes pairwise log-likelihoods $LL_p(x)$ for all $(x,p)$ with $x,p \in \test$ only once, and reuses them across iterations to update prefix scores without additional LLM inferences. 
Under the standard assumption that Transformer inference scales quadratically with sequence length, this results in a time complexity of $O(D^2 L^2)$, where $D = |\test|$ and $L$ is the average token length. 
Notably, \ours{} does not introduce tunable hyperparameters, whereas Min-K\% and Min-K\%++ require tuning $K$, and ReCaLL requires tuning the prefix size $n$, often without access to validation data.

Most baselines other than ReCaLL (Loss, Ref, Zlib, Min-K\%, and Min-K\%++) only compute the log-likelihood of each target text, yielding a time complexity of $O(D L^2)$. 
ReCaLL, in contrast, conditions on a prefix consisting of $n$ non-member examples, resulting in a complexity of $O(D (nL)^2) = O(n^2 D L^2)$. 
In practice, ReCaLL further sweeps over multiple values of $n$ to select the best configuration, which increases the overall cost to $O((1^2 + 2^2 + \cdots + n^2) D L^2) = O(n^3 D L^2)$. 
Their experiments also report using relatively large prefix sizes (e.g., $n=28$ in Figure~3 and Table~7) to achieve better performance. 
Although \ours{} scales quadratically with $D$ in theory, its overall computational cost for typical settings (e.g., $D \approx 1000$) is comparable to ReCaLL when accounting for this additional factor.

In addition, ReCaLL incurs an $O(n^2)$ memory overhead compared to other methods, including \ours{}, which can limit its applicability on memory-constrained hardware. 
By contrast, \ours{} is more amenable to parallelization and batching, and further efficiency improvements remain possible, for example, by approximating prefix scores using a subset of the test data.

\section{Conclusion}

We propose \ours{}, a membership inference method for large language models that jointly estimates membership scores and prompt effectiveness through an expectation-maximization procedure.  
Unlike prior work that relies on labeled non-members or assumes prompt quality in advance, \ours{} operates in a fully unsupervised gray-box setting, making it suitable for more realistic deployment scenarios.
Our method outperforms ReCaLL, even without its strong assumptions, and achieves state-of-the-art results on WikiMIA.  
\ours{} is modular and flexible, allowing different initialization strategies, scoring rules, and convergence criteria depending on the application context.

To support more rigorous and controlled evaluation, we introduce \ourbenchmark{}, a new benchmark built from the OLMo pretraining pipeline that allows fine-grained control over distributional overlap between members and non-members.  
Through comprehensive experiments, we show that \ours{} is robust across a wide range of difficulty settings, while also identifying scenarios where all existing methods struggle, particularly when member and non-member distributions are nearly identical.
Our findings highlight the importance of evaluating MIA methods under diverse and ambiguous conditions, and suggest that future progress will require methods that adapt to both prompt variability and fine-grained data overlap.

\newpage
\section*{Limitations}
Our paper focuses on detecting LLMs' pre-training data with the gray-box access where computing the probability of a text from output logits is possible.
However, many proprietary LLMs are usually further fine-tuned~\citep{ouyang2022training, chung2024scaling}, and they only provide generation outputs, which is the black-box setting.
We left the extension of our approach to MIAs for fine-tuned LLMs~\citep{song2019auditing, jagannatha2021membership, mahloujifar2021membership, shejwalkar2021membership, mireshghallah2022quantifying, tu2024dice, feng2024exposing} or LLMs with black-box access~\citep{dong2024generalization, zhou2024dpdllm, kaneko2024sampling} as future work.


\bibliography{reference}

\clearpage
\appendix
\section{Metrics for Prefix Scores}
\label{sec:prefix_scores_metrics}

\begin{figure}[t!]
\centering

\includegraphics[width=0.44\textwidth]{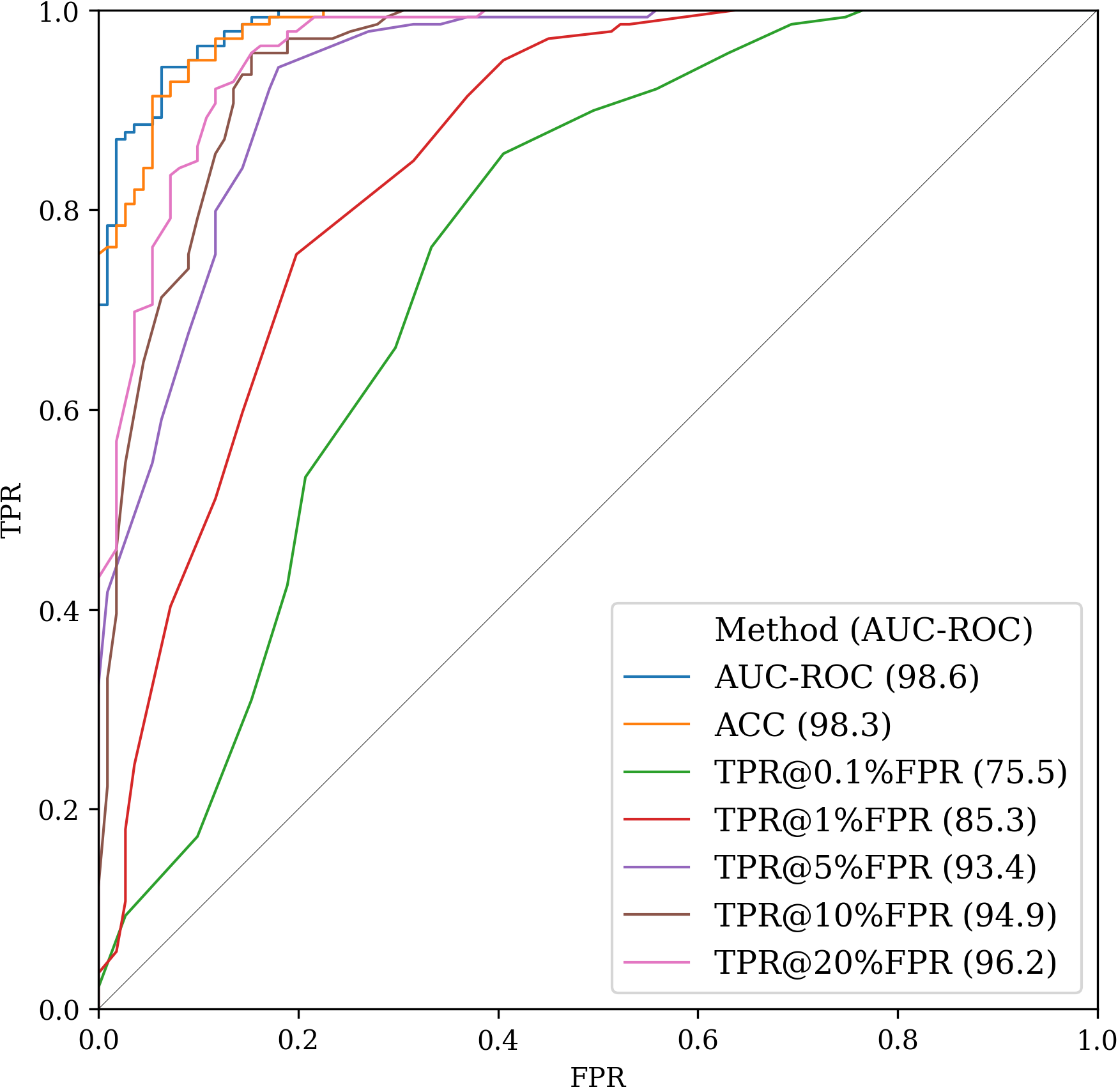}

\caption{
\label{fig:prefix_scores_metrics} 
ROC curves for MIA using the negative prefix score as the membership score, evaluated with different metrics for prefix scores in the oracle setting on WikiMIA~\citep{shi2023detecting} (length 128) using Pythia-6.9B~\citep{biderman2023pythia}.
}
\end{figure}

Figure~\ref{fig:prefix_scores_metrics} shows ROC curves when negative prefix scores, computed using different metrics, are used directly as membership scores.  
We compare prefix scoring metrics including AUC-ROC, Accuracy, and TPR@$k$\%FPR for $k \in \{0.1, 1, 5, 10, 20\}$.  
Among them, using AUC-ROC to compute prefix scores yields the best result, achieving 98.6\% AUC-ROC for membership inference.

\section{Comparison with ReCaLL}
\label{sec:recall_unfairness}

As explained in \S\ref{sec:recall}, the original ReCaLL~\citep{xie2024recall} uses labeled data from the test dataset, which is unfair to compare with the above baselines and ours.
More precisely, $p_i$ in the prefix $p = p_1 \oplus p_2 \oplus \cdots \oplus p_n$ are known non-members from the test set $\test$, and they are excluded from the test dataset for evaluation, i.e., $\test' = \test \setminus \{ p_1, p_2, \cdots, p_n \}$.
However, we measure the performance of ReCaLL with different prefix selection methods to understand how ReCaLL is sensitive to the prefix choice and use it as a reference instead of a direct fair comparison.

Since changing the test dataset every time for different prefixes does not make sense and makes the comparison even more complicated, we keep them in the test dataset. 
A language model tends to repeat, so $\mathrm{LL}(p_i | p; \lm) \simeq 0$.
Because $\mathrm{LL}(p_i | p; \lm) \ll 0$, $\recall_{p}(p_i; \lm) \simeq 0$.
It is likely to $\recall_{p}(p_i; \lm) \ll \recall_{p}(x; \lm)$ for $x \in \test \setminus \{ p_1, p_2, \cdots, p_n \}$, meaning that ReCaLL will classify $p_i$ as a non-member.
The effect would be marginal if $|\test| \gg n$.
Otherwise, we should consider this when we read numbers in the result table.

The original ReCaLL~\citep{xie2024recall} is similar to \randnm, except they report the best score after trying all different $n$ values, which is again unfair.
The number of shots $n$ is an important hyper-parameter determining performance. 
A larger $n$ generally leads to a better MIA performance but increases computational cost with a longer $p$.

\section{TPR@1\%FPR Results}
\label{sec:tpr_at_low_fpr}

\begin{table*}[!th]
\begin{center}
\setlength\tabcolsep{2pt}
\scriptsize
\begin{tabular}{lccccccccccccccccccccc}
\toprule
{\bf  Method} & \multicolumn{3}{c}{\bf  \makecell{Mamba-1.4B}} & \multicolumn{3}{c}{\bf  \makecell{Pythia-6.9B}} & \multicolumn{3}{c}{\bf  \makecell{LLaMA-13B}} & \multicolumn{3}{c}{\bf  \makecell{NeoX-20B}} & \multicolumn{3}{c}{\bf  \makecell{LLaMA-30B}} & \multicolumn{3}{c}{\bf  \makecell{OPT-66B}} & \multicolumn{3}{c}{\bf  Average} \\ 
\cmidrule(lr){2-4} \cmidrule{5-7} \cmidrule(lr){8-10} \cmidrule(lr){11-13} \cmidrule(lr){14-16} \cmidrule(lr){17-19} \cmidrule(lr){20-22}
{\bf } & 32 & 64 & 128 & 32 & 64 & 128 & 32 & 64 & 128 & 32 & 64 & 128 & 32 & 64 & 128 & 32 & 64 & 128 & 32 & 64 & 128 \\
\midrule
Loss      &  4.7 &  2.1 &  1.4 &  6.2 &  2.8 &  3.6 &  4.7 &  4.2 &  7.9 & 10.3 &  3.5 &  4.3 &  4.1 &  5.3 &  7.2 &  6.5 &  3.5 &  3.6 &  6.1 &  3.6 &  4.7 \\
Ref       &  0.5 &  0.7 &  0.7 &  1.6 &  1.1 &  1.4 &  2.3 &  3.9 &  2.9 &  3.1 &  2.5 &  1.4 &  1.3 &  2.5 &  3.6 &  1.8 &  1.8 &  0.7 &  1.8 &  2.1 &  1.8 \\
Zlib      &  4.1 &  4.9 &  7.2 &  4.9 &  6.0 & 11.5 &  5.7 &  8.1 & 12.9 &  9.3 &  6.3 &  5.0 &  4.9 &  9.5 & 10.1 &  5.7 &  7.0 & 11.5 &  5.8 &  7.0 &  9.7 \\
Min-K\%   &  7.0 &  4.2 &  5.8 &  8.8 &  3.9 &  7.2 &  5.2 &  6.0 & 15.1 & 10.6 &  3.9 &  7.2 &  4.7 &  7.0 &  5.8 &  9.0 &  7.7 &  8.6 &  7.5 &  5.5 &  8.3 \\
Min-K\%++ &  4.1 &  7.0 &  1.4 &  5.9 & 10.6 & 10.1 & 10.3 & 12.0 & 25.2 &  6.2 &  9.5 &  1.4 &  8.3 &  6.7 &  9.4 &  3.6 & 12.0 & 13.7 &  6.4 &  9.6 & 10.2 \\
\cmidrule{1-22}
\rowcolor{ReCaLLColor}
\avg      &  3.9 &  0.4 &  5.0 &  8.0 &  1.1 &  7.9 &  3.1 &  7.0 &  6.5 &  6.2 &  2.1 &  8.6 &  2.8 &  6.7 &  8.6 &  2.6 &  2.1 &  4.3 &  4.4 &  3.2 &  6.8 \\
\rowcolor{ReCaLLColor}
\avgp     &  0.5 &  0.4 &  0.7 &  1.8 &  0.4 &  0.0 &  0.0 &  0.7 &  0.0 &  1.3 &  0.7 &  0.0 &  0.0 &  0.0 &  2.9 &  2.1 & 12.3 & 24.5 &  0.9 &  2.4 &  4.7 \\
\rowcolor{ReCaLLColor}
\randm    &  0.8 &  0.1 &  0.6 &  0.9 &  0.0 &  1.9 &  0.2 &  0.4 &  7.6 &  0.5 &  0.3 &  1.6 &  0.4 &  0.6 &  8.1 &  0.7 &  0.1 &  0.9 &  0.6 &  0.2 &  3.4 \\
\rowcolor{ReCaLLColor}
\rand      &  3.7 &  3.9 &  2.4 &  2.3 &  3.2 &  7.6 &  1.6 &  2.7 &  7.3 &  4.4 &  5.0 &  4.7 &  1.6 &  3.2 &  7.9 &  2.1 &  3.2 &  3.2 &  2.6 &  3.5 &  5.5 \\
\rowcolor{ReCaLLColor}
\randnm   & 19.2 &  8.3 & 15.4 & 12.6 & 10.5 & 18.7 & 18.5 & 17.2 &  7.5 & 12.9 & 11.6 & 12.5 & 13.8 & 18.7 &  8.1 &  5.0 &  5.0 &  6.6 & 13.7 & 11.9 & 11.5 \\
\rowcolor{ReCaLLColor}
\toppref   & 12.7 &  4.2 & 25.2 & 16.0 &  1.4 & 29.5 & 14.2 &  9.2 &  7.9 & 13.4 & 13.7 & 20.9 & 27.1 & 29.9 &  8.6 &  3.9 &  5.6 &  9.4 & 14.6 & 10.7 & 16.9 \\
\rowcolor{ReCaLLColor}
\citet{xie2024recall} & 11.2 & 11.0 &  4.0 & 28.5 & 20.7 & 33.3 & 13.3 & 30.1 & 26.3 & 25.3 &  6.9 & 30.3 & 18.4 & 18.3 &  1.0 &  8.3 &  5.3 &  6.1 & 17.5 & 15.4 & 16.9 \\
\cmidrule{1-22}
\rowcolor{OursColor}
\ours     & 54.0 & 47.9 & 51.8 & 50.4 & 56.0 & 47.5 & 66.4 & 75.7 & 58.3 & 51.4 & 64.1 & 59.0 & 61.5 & 66.2 & 71.9 & 83.5 & 73.2 & 39.6 & 61.2 & 63.8 & 54.7 \\
\bottomrule
\end{tabular}
\end{center}
\caption{TPR@1\%FPR results on WikiMIA benchmark. The second block (grey) is the ReCaLL-based baselines. \randm, \randnm, ReCaLL, and \toppref~use labels in the test dataset, so comparing them with others is unfair. We report their scores for reference. We borrow the original ReCaLL results from \citet{xie2024recall}, which is also unfair to be compared with ours and other baselines.}
\label{tab:wikimia_tpr_at_low_fpr}

\end{table*}

\begin{table*}[!th]
\begin{center}
\scriptsize
\begin{tabular}{lcccccccccccc}
\toprule
{\bf  Method} & \multicolumn{2}{c}{\bf  \makecell{Easy}} & \multicolumn{2}{c}{\bf  \makecell{Medium}} & \multicolumn{2}{c}{\bf  \makecell{Hard}} & \multicolumn{2}{c}{\bf  \makecell{Random}} & \multicolumn{2}{c}{\bf  \makecell{Mix-1}} & \multicolumn{2}{c}{\bf  \makecell{Mix-2}} \\ 
\cmidrule(lr){2-3} \cmidrule{4-5} \cmidrule(lr){6-7} \cmidrule(lr){8-9} \cmidrule(lr){10-11} \cmidrule(lr){12-13}
{\bf } & 64 & 128 & 64 & 128 & 64 & 128 & 64 & 128 & 64 & 128 & 64 & 128 \\
\midrule
Loss      &  2.8 & 12.8 &  7.2 &  1.4 &  0.1 &  1.2 &  1.3 &  0.7 &  7.2 &  1.7 &  0.0 &  0.7 \\
Ref       &  6.2 &  4.0 &  4.9 &  0.6 &  1.0 &  0.9 &  1.2 &  1.2 &  8.4 &  0.5 &  0.2 &  1.6 \\
Zlib      &  2.0 &  9.8 &  6.7 &  1.1 &  0.2 &  1.6 &  0.9 &  0.7 &  6.4 &  1.7 &  0.0 &  0.7 \\
Min-K\%   &  1.3 &  6.5 &  5.8 &  1.4 &  0.1 &  1.3 &  1.1 &  0.7 &  6.1 &  2.0 &  0.0 &  0.7 \\
Min-K\%++ &  1.4 &  8.0 &  5.0 &  0.7 &  0.4 &  1.0 &  1.0 &  0.4 &  5.0 &  0.9 &  0.0 &  0.5 \\
\cmidrule{1-13}
\rowcolor{ReCaLLColor}
\avg      &  4.1 & 11.5 &  4.0 &  1.7 &  0.2 &  2.2 &  1.2 &  0.6 &  6.1 &  2.2 &  0.0 &  0.9 \\
\rowcolor{ReCaLLColor}
\avgp     & 11.7 &  0.1 &  2.6 &  7.2 &  0.7 &  1.6 &  0.7 &  1.4 &  4.8 & 12.1 &  0.1 &  0.0 \\
\rowcolor{ReCaLLColor}
\randm    &  3.0 &  4.9 &  2.4 &  1.1 &  0.4 &  2.2 &  0.9 &  0.8 &  7.6 &  1.3 &  0.0 &  0.4 \\
\rowcolor{ReCaLLColor}
\rand     &  4.3 &  7.8 &  3.7 &  1.7 &  0.4 &  2.7 &  1.0 &  0.8 & 10.6 &  3.0 &  0.0 &  0.7 \\
\rowcolor{ReCaLLColor}
\randnm   & 16.9 & 14.2 &  5.2 &  1.8 &  0.3 &  1.9 &  1.0 &  0.8 &  9.2 &  2.9 &  0.0 &  1.1 \\
\rowcolor{ReCaLLColor}
\toppref  & 22.0 & 16.6 &  6.3 &  1.9 &  0.4 &  2.2 &  1.1 &  1.4 &  8.1 &  5.1 &  0.0 &  0.5 \\
\cmidrule{1-13}
\rowcolor{OursColor}
\ours     & 95.0 & 52.1 & 79.8 & 96.7 &  1.8 &  1.0 &  1.1 &  1.4 & 12.2 &  3.8 & 14.8 &  4.3 \\
\bottomrule
\end{tabular}
\end{center}
\caption{TPR@1\%FPR results on \ourbenchmark~benchmark. The second block (grey) is the ReCaLL-based baselines. \randm, \randnm, ReCaLL, and \toppref~use labels in the test dataset, so comparing them with others is unfair. We report their scores for reference.}
\label{tab:olmomia_tpr_at_low_fpr}

\end{table*}

TPR@low FPR is a useful MIA evaluation metric~\citep{carlini2022membership} in addition to AUC-ROC, especially when developing a new MIA and comparing it with other MIAs.
Due to the space limitation in the main text, we put TPR@low FPR here: Table~\ref{tab:wikimia_tpr_at_low_fpr} for WikiMIA and Table~\ref{tab:olmomia_tpr_at_low_fpr} for \ourbenchmark.

\end{document}